\documentclass[aps,pre,reprint,groupedaddress, floatfix]{revtex4-2}

\usepackage{amsmath,amssymb}
\usepackage{newtxtext,newtxmath}
\usepackage{algorithm}
\usepackage{algpseudocode}
\usepackage{bm,color}

\usepackage{graphicx,graphics}

\algnewcommand{\Initialize}[1]{%
  \State \textbf{Initialize:}
  \State \hspace*{\algorithmicindent}\parbox[t]{0.8\linewidth}{\raggedright #1}
}

\newcommand{\Fref}[1]{Fig.\ref{#1}}
\newcommand{\eref}[1]{eq.(\ref{#1})}

\begin{document}

\title{Active pooling design in group testing
based on Bayesian posterior prediction}

\author{Ayaka Sakata}
\email[]{ayaka@ism.ac.jp}
\affiliation{Institute of Statistical Mathematics, 
10-3 Midori-cho, Tachikawa, Tokyo 190-8562, Japan}
\affiliation{Department of Statistical Science, 
The Graduate University for Advanced Science
(SOKENDAI), Hayama-cho, Kanagawa 240-0193, Japan}
\affiliation{JST PRESTO, 4-1-8 Honcho, Kawaguchi, Saitama, 332-0012, Japan}

\date{\today}

\begin{abstract}
For identifying infected patients in a population,
group testing is an effective method to reduce the number of tests and
correct test errors.
In group testing, 
tests are performed on pools of specimens collected from patients,
where the number of pools is lower than that of patients.
The performance of group testing considerably depends on the 
design of pools and algorithms that are used for inferring the
infected patients from the test outcomes.
In this paper, an adaptive design method of pools 
based on the predictive distribution is proposed
in the framework of Bayesian inference.
The proposed method executed using a belief propagation algorithm 
results in more accurate identification of the infected patients,
compared with 
the group testing performed on random pools
determined in advance.
\end{abstract}


\maketitle

\section{Introduction}

Identification of infected patients from a large population
using clinical tests, such as blood tests and polymerase chain reaction tests,
requires significant operating costs.
Group testing is one of the approaches 
to reduce such costs
by performing tests on pools of specimens
obtained from patients \cite{Dorfman,GT_book}.
When the fraction of infected patients 
in a population is sufficiently small, 
the infected patients can be identified 
from tests on pools whose number is smaller than that of the patients.
Originally, group testing was developed for blood testing by Dorfman
and is now applied to various fields, 
such as quality control in product testing \cite{product}
and multiple access communication \cite{Wolf}.

Group testing is roughly classified into non-adaptive 
and adaptive.
In non-adaptive group testing,
all pools are determined in advance and fixed during all tests. 
In adaptive group testing, pools are designed
sequentially, depending on the previous test outcomes.
Dorfman's original study considered the simplest adaptive procedure, 
the so-called two-stage testing; here,
in the first round, tests are performed on pools designed in advance,
and all patients belonging to the positive pool are individually tested in the subsequent stage.
A generalization of the two-stage testing is known as a binary splitting method \cite{Sobel1,Sobel2},
where the positive pool in the previous stage 
is split into two subpools. Tests in the subsequent stage are performed on the subpools
until the infected patients are identified.
Further, the splitting of the positive pools into several subsets larger than two sometimes 
reduces the number of tests required for identifying the infected patients \cite{Hwang}.
These splitting-based methods 
are effective when the number of infected patients is sufficiently small.
However, the splitting-based methods 
exhibit a limitation in the correction of false negative results
because patients in the negative pools are never tested again,
even when the negative result is false.

Different from the splitting-based design,
active design of data sampling has been studied in statistics
and machine learning, known as experiments design \cite{Fedorov,Pukelsheim}, 
active learning \cite{Cohn,Settles}, and Bayesian optimization \cite{Brochu,Shahriari}.
In these approaches, the optimal method to select training data for efficient learning
is developed considering several criteria that 
quantify informativeness of the unknown data.
The active design of data sampling improves the performance of 
algorithms in several fields, 
such as text classification \cite{Cohn}, semi-supervised learning \cite{Zhu}, 
and support vector machine \cite{Tong}.
Active data sampling 
is particularly effective when data possess uncertainty due to a noisy 
generative process and there exists a limitation in the number of data sampling.
In the context of group testing,
active sampling of data corresponds to the active design of pools for the subsequent stage.
Further, noise is observed during tests and the number of tests should be reduced; active sampling makes a significant contribution by addressing these issues.


In this paper, we propose an active pooling design method 
employing Bayesian inference
for efficient identification of infected patients using group testing.
Bayesian modeling can consider the 
finite false probabilities in the test
and provide a measure to quantify the uncertainty, posterior predictive distribution.
We sequentially design pools based on the predictive distribution
in adaptive group testing.
The procedure is executed using 
a statistical-physics-based algorithm, belief propagation (BP) 
\cite{Mezard-Montanari,Mezard_GT,Johnson,Kanamori},
which achieves a reasonable approximation of estimates
with a feasible computational cost \cite{Sakata}.
We demonstrate that, compared with the approach that uses randomly generated pools, the proposed pooling method 
effectively corrects errors with a smaller number of tests.

\section{Mathematical formulation}

Let us denote the true state of $N$-patients by $\bm{X}^{(0)}\in\{0,1\}^N$,
where $X_i^{(0)}=1$ and $X_i^{(0)}=0$ 
indicate that the $i$-th patient is infected and not infected,
respectively.
The pooling of the patients is determined by
a matrix $\bm{F}\in\{0,1\}^{M\times N}$,
where $M(<N)$ is the number of pools and $F_{\mu i}=1$ and $F_{\mu i}=0$
indicate that the $i$-th patient is in the $\mu$-th pool
and is not, respectively.
The true state of the $\mu$-th group, denoted by $T(\bm{X}^{(0)},\tilde{\bm{F}}_\mu)$,
where $\tilde{\bm{F}}_\mu$ is the $\mu$-th row vector of $\bm{F}$,
is given by $T(\bm{X}^{(0)},\tilde{\bm{F}}_\mu)=\mathop{\vee}_{i=1}^NF_{\mu i}X_i^{(0)}$,
where $\vee_{i=1}^N f_i=f_1\vee f_2\vee \cdots\vee f_N$
denotes the logical sum of $N$ components.
Namely, when the $\mu$-th pool contains 
at least one infected patient,
the state of the $\mu$-th pool is 1 (positive); otherwise,
it is 0 (negative).

The test error is modeled by a function $C(\cdot)$ 
that returns 0 or 1 according to the probability conditioned by the input as 
\begin{align}
P(C(a)=1|a=1)&=p_{\mathrm{TP}},~~~P(C(a)=0|a=1)=1-p_{\mathrm{TP}}\\
P(C(a)=1|a=0)&=p_{\mathrm{FP}},~~~P(C(a)=0|a=0)=1-p_{\mathrm{FP}},
\end{align}
and
$p_{\mathrm{TP}}$ and $p_{\mathrm{FP}}$ correspond to the 
true-positive (TP) and false-positive (FP) probabilities
in the test, respectively \cite{Johnson, Sakata}.
We assume that the test errors are independent of each other; further,
from the property of $C(\cdot)$,
the generative model of $\bm{Y}$
is given by $P_{\mathrm{gen}}(\bm{Y}|\bm{X}^{(0)},\bm{F})
=\prod_{\mu=1}^M P_{\mathrm{gen}}(Y_\mu|\bm{X}^{(0)},\tilde{\bm{F}}_\mu)$,
where
\begin{align}
\nonumber
    P_{\mathrm{gen}}(Y_\mu &|\bm{X}^{(0)},\tilde{\bm{F}}_\mu)=\{p_{\mathrm{TP}}
    Y_\mu + (1-p_{\mathrm{TP}})(1-Y_\mu)\}T(\bm{X}^{(0)},\tilde{\bm{F}}_\mu)\\
    &+\{p_{\mathrm{FP}}Y_\mu+(1-p_{\mathrm{FP}})(1-Y_\mu)\}
    (1-T(\bm{X}^{(0)},\tilde{\bm{F}}_\mu))
    \label{eq:Y_generate}
\end{align}
is a Bernoulli distribution conditioned by $\bm{X}$ and $\bm{F}$.

Currently, we aim to infer the true states of patients $\bm{X}^{(0)}$
from the observation $\bm{Y}$.
To this end, Bayes formula is considered.
Further, we introduce the prior distribution of the patient states 
$P_{\mathrm{pri}}(X_i)\sim\mathrm{Bernoulli}(\rho)$,
where $\rho\in[0,1]$ is the assumed infection probability.
Following the Bayes rule, the
posterior distribution is given by 
$P_{\mathrm{post}}(\bm{X}|\bm{Y})\propto P_{\mathrm{gen}}(\bm{Y}|\bm{X})\prod_iP_{\mathrm{pri}}(X_i|\rho)$.
The $i$-th patient's state
is identified on the basis of the 
marginal distribution $P_{\mathrm{post}}(X_i|\bm{Y})=\sum_{\bm{X}\backslash X_i}P_{\mathrm{post}}(\bm{X}|\bm{Y})$,
where $\bm{X}\backslash X_i$ denotes the components of $\bm{X}$
other than $X_i$.
As the variable $X_i$ is binary, we can represent the 
marginal distribution using a Bernoulli
probability $\theta_i$ as
\begin{align}
    P_{\mathrm{post}}(X_i|\bm{Y})=\theta_i(\bm{Y})X_i+(1-\theta_i(\bm{Y}))(1-X_i),
\end{align}
and $\theta_i(\bm{Y})$ corresponds to the infection probability
estimated under the test result $\bm{Y}$,
namely, the probability that $X_i=1$.
We have to convert the returned probability to a binary value for the 
identification of the patients' states.
The simplest estimate of $X_i^{(0)}$ is
the maximum a posteriori (MAP) estimator given by
\begin{align}
    X_i^{(\mathrm{MAP})}=\mathbb{I}(\theta_i>0.5),
\end{align}
where $\mathbb{I}(a)$ is the indicator function
whose value is 1 when $a$ is true, and 0 otherwise.

\section{Adaptive design of pools}

Here, we divide $M$-tests into 
$M_{\mathrm{ini}}$-tests under pools fixed in advance as the initial stage
and 
$M_{\mathrm{ada}}$-tests sequentially performed on
actively designed pools as the adaptive stage. 
Hence, $M=M_{\mathrm{ini}}+M_{\mathrm{ada}}$. 
We denote the index set of patients who
are in the $\nu$-th pool as $\pi(\nu)$,
where $F_{\nu i}=1$ for $i\in\pi(\nu)$; otherwise, 0.
We consider the determination of $\pi(\nu+1)$ $(\nu\geq M_{\mathrm{ini}})$ 
among possible pools denoted by ${\cal P}$ based on the
$1,\cdots,\nu$-th test outcomes, denoted by $\bm{Y}_{(\nu)}=[Y_1,Y_2,...,Y_\nu]^{\mathrm{T}}$,
which are performed on pools $\pi(1),\cdots,\pi(\nu)$.
The predictive distribution for the unknown result of the test $Y\in\{0,1\}$,
which will be performed on a certain pool $\pi\in{\cal P}$, is defined as
\begin{align}
P_{\mathrm{pre}}(Y|\bm{Y}_{(\nu)},\pi)=\sum_{\bm{X}_\pi}P_{\mathrm{gen}}(Y|\bm{X}_{\pi})P_{\mathrm{post}}(\bm{X}_{\pi}|\bm{Y}_{(\nu)}),
\end{align}
where $\bm{X}_{\pi}=\{X_i|i\in\pi\}$ and
$P_{\mathrm{post}}(\bm{X}_{\pi}|\bm{Y}_{(\nu)})=\sum_{\bm{X}\backslash\bm{X}_{\pi}}P_{\mathrm{post}}(\bm{X}|\bm{Y}_{(\nu)})$.
By setting $P_{\mathrm{post}}(\bm{X}_{\pi}=\bm{0}|\bm{Y}_{(\nu)})=q(\pi;\bm{Y}_{(\nu)})$,
which is the estimated probability under given $\bm{Y}_{(\nu)}$
that all patients in the pool $\pi$ are not infected,
the predictive distribution is expressed as
\begin{align}
\nonumber
P_{\mathrm{pre}}(Y|\bm{Y}_{(\nu)}&,\pi)=\{p_{\mathrm{TP}}Y+(1-p_{\mathrm{TP}})(1-Y)\}(1-q(\pi,\bm{Y}_{(\nu)}))\\
&+\{p_{\mathrm{FP}}Y+(1-p_{\mathrm{FP}})(1-Y)\}q(\pi,\bm{Y}_{(\nu)}).
\label{eq:predictive_dist}
\end{align}
The predictive distribution measures 
the adequacy of the posterior distribution to describe the unknown data,
and is used as a modeling criterion in Bayesian inference \cite{Kitagawa}.
We use the predictive distribution for active design of pools.
For intuitive discussion,
let us consider the case that $P_{\mathrm{pre}}(Y=1|\bm{Y}_{(\nu)},\pi)$ and 
$P_{\mathrm{pre}}(Y=0|\bm{Y}_{(\nu)},\pi)$ are significantly different.
 We consider the case that 
they are close to 1 and close to 0, respectively.
This means that the posterior distribution
is consistent with the new observation
performed on the pool $\pi$
in the sense that 
the current posterior matches the new test result $Y=1$,
and $Y=0$ is supposed to be the test error. 
We do not consider this `explainable pool' in the subsequent stage 
because the test performed on the explainable pool is
not expected to modify the current posterior to be realistic. 
Instead, we consider the pool $\pi$ that gives comparable
$P_{\mathrm{pre}}(Y=1|\bm{Y}_{(\nu)},\pi)$ and $P_{\mathrm{pre}}(Y=0|\bm{Y}_{(\nu)},\pi)$,
where the posterior at step $\nu$ cannot explain the test result
performed on the pool $\pi$,
and hence, the test result is expected to correct the posterior to explain it.

This strategy can be expressed by the
maximization of the predictive entropy at step $\nu+1$ defined as
\begin{align}
S(\bm{Y}_{(\nu)},\pi)=-\sum_{Y}P_{\mathrm{pre}}(Y|\bm{Y}_{(\nu)},\pi)\ln P_{\mathrm{pre}}(Y|\bm{Y}_{(\nu)},\pi),
\label{eq:S_pre}
\end{align}
where $P_{\mathrm{pre}}(Y=1|\bm{Y}_{(\nu)},\pi)=P_{\mathrm{pre}}(Y=0|\bm{Y}_{(\nu)},\pi)=0.5$
gives the entropy maximum.
Active design of data sampling based on the 
entropy maximization is known as uncertainty sampling 
in active learning \cite{entropy_max,entropy_max2}.
As shown in \eref{eq:predictive_dist}, 
the predictive entropy is expressed by one parameter $q(\pi,\bm{Y}_{(\nu)})$.
Regarding the predictive entropy as a function of $q\in[0,1]$,
the maximum of the predictive entropy is achieved at $q=q^{*}$ given by
\begin{align}
q^*=\frac{p_{\mathrm{TP}}-0.5}{p_{\mathrm{TP}}-p_{\mathrm{FP}}},
\label{eq:def_q}
\end{align}
where $p_{\mathrm{FP}}<0.5\leq p_{\mathrm{TP}}$ is assumed.
We determine the $\nu+1$-th pool as
\begin{align}
\pi(\nu+1)=\arg\min_{\pi\in {\cal P}}|q(\pi,\bm{Y}^{(\nu)})-q^*|.
\end{align}

The remaining task is the calculation of $q(\pi;\bm{Y}_{(\nu)})$
for possible $\pi$ under the given test results $\bm{Y}_{(\nu)}$.
The mathematical form of $q(\pi;\bm{Y}_{(\nu)})$
depends on the size of $\pi$, denoted by $|\pi|$.
When $|\pi|=1$,
we obtain 
\begin{align}
q(\pi,\bm{Y}_{(\nu)})=1-\theta_{\pi}(\bm{Y}_{(\nu)}).
\end{align}
For larger pools, the correlation between the patients in the pool 
should be considered for the exact evaluation of $q(\pi;\bm{Y}_{(\nu)})$.
For example, when $\pi=\{i,j\}$ $(|\pi|=2)$,
we obtain
\begin{align}
q(\pi,\bm{Y}_{(\nu)})=\chi_{ij}(\bm{Y}_{(\nu)})+
(1-\theta_{i}(\bm{Y}_{(\nu)}))(1-\theta_{j}(\bm{Y}_{(\nu)})),
\label{eq:q_pi2}
\end{align}
where $\chi_{ij}(\bm{Y}_{(\nu)})=E_{\mathrm{post}|\bm{Y}_{(\nu)}}[X_iX_j]-\theta_i(\bm{Y}_{(\nu)})\theta_j(\bm{Y}_{(\nu)})$ is the susceptibility
and $E_{\mathrm{post}|\bm{Y}_{(\nu)}}[\cdot]$ denotes the average according to the 
posterior distribution $P_{\mathrm{post}}(\bm{X}|\bm{Y}_{(\nu)})$.

Next, we discuss the relationship between $q^*$, $p_{\mathrm{TP}}$, and $p_{\mathrm{FP}}$.
From the definition of $q^*$, \eref{eq:def_q},
if $p_{\mathrm{TP}}>1-p_{\mathrm{FP}}$, then $q^*<0.5$.
This indicates that 
the pools with $q(\pi,\bm{Y}_{(\nu)})<0.5$ are likely to be chosen when 
$p_{\mathrm{TP}}>1-p_{\mathrm{FP}}$.
In other words, when the probability that at least one patient in a pool is infected is 
larger than the probability that no one is infected,
the pool tends to be chosen.
This can be understood as follows.
Introducing false negative probability $p_{\mathrm{FN}}$,
$p_{\mathrm{TP}}>1-p_{\mathrm{FP}}$ is equivalent to $p_{\mathrm{FN}}<p_{\mathrm{FP}}$.
This means that false test results are mainly contained in positive results. 
Hence, pools with $q(\pi,\bm{Y}_{(\nu)})<0.5$
contain significant uncertainty, compared with $q(\pi,\bm{Y}_{(\nu)})>0.5$.
Therefore, in the active pooling design based on uncertainty,
pools with $q(\pi,\bm{Y}_{(\nu)})<0.5$ are preferably chosen, 
when $p_{\mathrm{TP}}>1-p_{\mathrm{FP}}$.
Following the same logic, 
we can understand that the pool with $q(\pi,\bm{Y}_{(\nu)})>0.5$ is likely to be chosen when
$p_{\mathrm{TP}}<1-p_{\mathrm{FP}}$ .

\section{Implementation by belief propagation}

The computation of the marginal distribution 
requires the exponential order of the sums, and thus is intractable.
We approximately calculate the marginal distribution using the BP
algorithm \cite{Mezard_GT,Johnson,Kanamori}.
Compared with the approximation using the BP algorithm with the exact calculation at a small size,
the BP algorithm has sufficient approximation performance
when applied to group testing \cite{Sakata}.
In this study, we use the BP algorithm as a reasonable method owing
to its approximation accuracy and computational time.
In Appendix \ref{sec:app_BP},
the BP algorithm for calculating the infection probability
given by the posterior distribution is summarized.
We denote the obtained estimates of $\theta_i$ and the 
corresponding MAP estimator as $\hat{\theta}_i$ and $\hat{X}_i^{(\mathrm{MAP})}=\mathbb{I}(\hat{\theta}_i>0.5)$,
respectively.
We measure the accuracy of the MAP estimator by 
the TP and FP rates given by
\begin{align}
    \mathrm{TP}=\frac{\sum_i X_i^{(0)}\hat{X}_i^{(\mathrm{MAP})}}{\sum_iX_i^{(0)}},~~~
    \mathrm{FP}=\frac{\sum_i(1-X_i^{(0)})\hat{X}^{(\mathrm{MAP})}_i}{\sum_i(1-X_i^{(0)})},
\end{align}
respectively.
A TP value larger than $p_{\mathrm{TP}}$ and an FP value smaller than 
$p_{\mathrm{FP}}$ indicate that 
the BP-based identification has
better performance than the parallel test of $N$-patients.

To apply the BP algorithm to an adaptive test,
we need to obtain $q(\pi,\bm{Y}_{(\nu)})$
for each $\nu$ $(>M_{\mathrm{ini}})$.
For its exact computation,
we need multibody correlations between patients
except when $|\pi|=1$,
although the BP algorithm returns one-body information.
In this study, we use the simplest approximation 
provided by the BP algorithm as 
$\hat{q}(\pi,\bm{Y}_{(\nu)})\equiv\prod_{i=1}^{|\pi|}(1-\hat{\theta}_{\pi_i})$,
where $\pi_i~(i=1,\cdots,|\pi|)$ is the $i$-th component in the pool $\pi$,
to avoid the increase in the computational time
required for the calculation of multibody correlation.
Further, to reduce the time of the 
computation of $q(\pi,\bm{Y}^{(\nu)})$ for all possible $\pi$, we focus on the subspace of pools
${\cal P}_1=\{\pi||\pi|=1,\pi\in{\cal P}\}$ and ${\cal P}_2=\{\pi||\pi|\leq 2,\pi\in{\cal P}\}$;
hence, ${\cal P}_1\subset {\cal P}_2\subset {\cal P}$.
In principle, BP can approximately compute the correlation between patients
by deriving conditional posterior expectations,
which requires additional computations of the order of $O(N!\slash(N-|\pi|)!)$
according to the product-rule of conditional joint distributions.
As an example, we calculate the susceptibility using the BP algorithm
and implement active pooling design on the basis of 
$q(\pi,\bm{Y}^{(\nu)})$ for $|\pi|=2$ case,
as shown in Appendix \ref{sec:app_correlation}.
The consideration of the susceptibility does not provide
large improvements in terms of TP and FP rates in the problem setting studied herein.
Hence, we use one-body approximation $\hat{q}(\pi,\bm{Y}_{(\nu)})$ throughout the study.

\begin{algorithm}[H]
\caption{Group testing with active pooling design using the belief propagation (BP)  algorithm}
\label{alg:adaptiveGT}
\begin{algorithmic}[1]
\Require {$\bm{Y}_{(M_{\mathrm{ini}})}\in\{0,1\}^{M_{\mathrm{ini}}}$ and $\bm{F}_{(M_{\mathrm{ini}})}\in\{0,1\}^{M_{\mathrm{ini}}\times N}$}
\Ensure {$\hat{\bm{\theta}}\in[0,1]^N$}
    \State{$\hat{\bm{\theta}}\gets \mathrm{BP}(\bm{Y}_{\mathrm{ini}},\bm{F}_{\mathrm{ini}})$}
    \For{$\nu=M_{\mathrm{ini}}+1 \, \ldots \, M$}
    	\State{$\hat{q}(\pi;\bm{Y}_{(\nu-1)})\gets\prod_{i=1}^{|\pi|}(1-\hat{\theta}_{\pi_i})$ for $\pi\in{\cal P}$}
	\State{$\pi(\nu)\gets\arg\min_{\pi\in{\cal P}}|\hat{q}(\pi;\bm{Y}_{(\nu-1)})-q^*|$}
	\State{$\tilde{\bm{F}}_\nu\gets[F_{\nu 1},\cdots,F_{\nu N}]$, where $F_{\nu i}=1$ for $i\in\pi(\nu)$, otherwise 0}
	\State{$Y_\nu\sim P_{\mathrm{gen}}(Y|\bm{X}^{(0)},\tilde{\bm{F}}_\nu)$}\Comment{Test result performed on $\pi(\nu)$}
	\State{$\bm{Y}_{(\nu)}\gets [\bm{Y}_{(\nu-1)};Y_{\nu}]$}
	\State{$\bm{F}_{(\nu)}\gets[\bm{F}_{(\nu-1)};\tilde{\bm{F}}_{\nu}]$}
	\State{$\hat{\bm{\theta}}\gets\mathrm{BP}(\bm{Y}_{(\nu)},\bm{F}_{(\nu)})$}
    \EndFor
\end{algorithmic}
\end{algorithm}

The setting of the numerical simulation described in this section  is as follows.
Let us denote the longitudinal coupling
of matrices or vectors 
$\bm{a}$ and $\bm{b}$ that have the same number of columns as $[\bm{a};\bm{b}]$.
Hence, $\bm{F}=[\tilde{\bm{F}}_1;\tilde{\bm{F}}_2;\cdots;\tilde{\bm{F}}_M]$.
The submatrix of $\bm{F}$ given by from the $1$-st to the $\nu$-th row vectors is 
denoted by
$\bm{F}_{(\nu)}=[\tilde{\bm{F}}_1;\cdots;\tilde{\bm{F}}_\nu]$;
hence, $\bm{F}_{(\nu+1)}=[\bm{F}_{(\nu)};\tilde{\bm{F}}_{\nu+1}]$.
The pooling matrix for the initial stage, $\bm{F}_{(M_{\mathrm{ini}})}$, is randomly generated 
under the constraint that the number of pools each patient belongs to
and the number of patients in each pool are fixed at 
$N_G(\ll N)$ and $N_O(\ll N)$, respectively.
Hence, $\sum_{i=1}^NF_{\mu i}=N_G$ for $\mu\leq M_{\mathrm{ini}}$ and 
$\sum_{\mu=1}^{M_{\mathrm{ini}}} F_{\mu i}=N_O~\forall i$ hold,
and the relationship $N_O=N_GM_{\mathrm{ini}}\slash N$ holds.
The corresponding test result
in the initial stage, $\bm{Y}_{(M_{\mathrm{ini}})}$,
is generated as $\bm{Y}_{(M_{\mathrm{ini}})}\sim P_{\mathrm{gen}}(\bm{Y}|\bm{X}^{(0)},\bm{F}_{(M_{\mathrm{ini}})})$.
The posterior distribution under given $\bm{Y}_{(M_{\mathrm{ini}})}$
and $\bm{F}_{(M_{\mathrm{ini}})}$
is approximately calculated using the BP algorithm.
For the subsequent adaptive stage,
we actively choose $\pi(M_{\mathrm{ini}}+1)$ among ${\cal P}_1$ or ${\cal P}_2$
based on the predictive entropy given by the posterior distribution of the initial stage.
Next, we construct $\tilde{\bm{F}}_{M_{\mathrm{ini}}+1}$, so that $F_{M_{\mathrm{ini}}+1,i}=1$ for $i\in\pi(M_{\mathrm{ini}}+1)$;
otherwise, 0.
The test result is generated as 
$Y_{M_{\mathrm{ini}}+1}\sim P_{\mathrm{gen}}(Y|\bm{X}^{(0)},\tilde{\bm{F}}_{M_{\mathrm{ini}}+1})$,
and we obtain the posterior distribution
under $\bm{F}_{(M_{\mathrm{ini}}+1)}=[\bm{F}_{(M_{\mathrm{ini}})};\tilde{\bm{F}}_{M_{\mathrm{ini}}+1}]$
and $\bm{Y}_{(M_{\mathrm{ini}}+1)}=[\bm{Y}_{(M_{\mathrm{ini}})};Y_{M_{\mathrm{ini}}+1}]$
using the BP algorithm.
This adaptive test procedure is repeated $M_{\mathrm{ada}}$-times,
where $M=M_{\mathrm{ini}}+M_{\mathrm{ada}}$, and 
the state of patients is determined by 
the MAP estimator corresponding to $\hat{\theta}(\bm{Y},\bm{F})$,
where $\bm{Y}=[\bm{Y}_{(M_{\mathrm{ini}})};Y_{M_{\mathrm{ini}+1}};\ldots;Y_M]$
and $\bm{F}=[\bm{F}_{(M_{\mathrm{ini}})};\tilde{\bm{F}}_{M_{\mathrm{ini}}+1};\ldots;\tilde{\bm{F}}_M]$.
The pseudocode is summarized in Algorithm \ref{alg:adaptiveGT},
where $\mathrm{BP}(\bm{Y},\bm{F})$ indicates the calculation of the 
infection probability using the BP algorithm under the input $\bm{Y}$ and $\bm{F}$
(see Appendix \ref{sec:app_BP}).

The true state of patients $\bm{X}^{(0)}$ is randomly generated
under the constraint that $\sum_iX_i^{(0)}=N\rho$.
Here, we assume that the correct parameters $\rho$,
$p_{\mathrm{TP}}$, and $p_{\mathrm{FP}}$ are known in advance.
For more general cases where the estimation of unknown parameters
is required, we can construct their estimators
by combining the BP algorithm with the expectation-maximization method, or 
introducing a hierarchical Bayes model \cite{Sakata}.


\begin{figure}
\begin{minipage}{0.49\hsize}
\centering
    \includegraphics[width=1.7in]{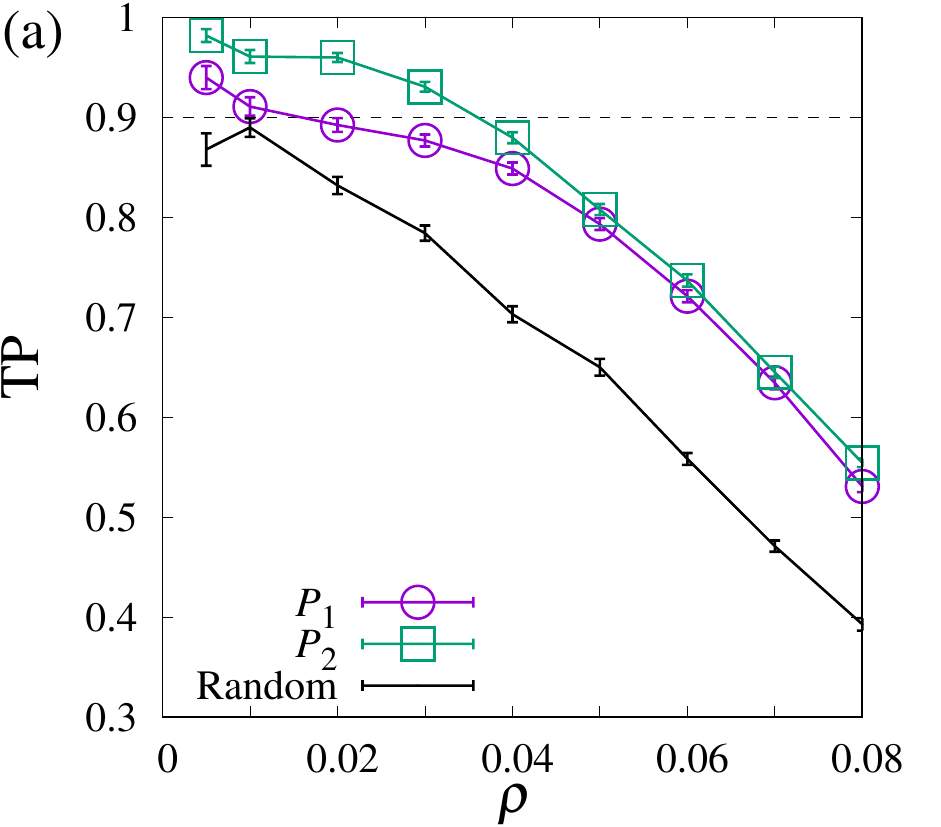}
\end{minipage}
\begin{minipage}{0.49\hsize}
\centering
\includegraphics[width=1.7in]{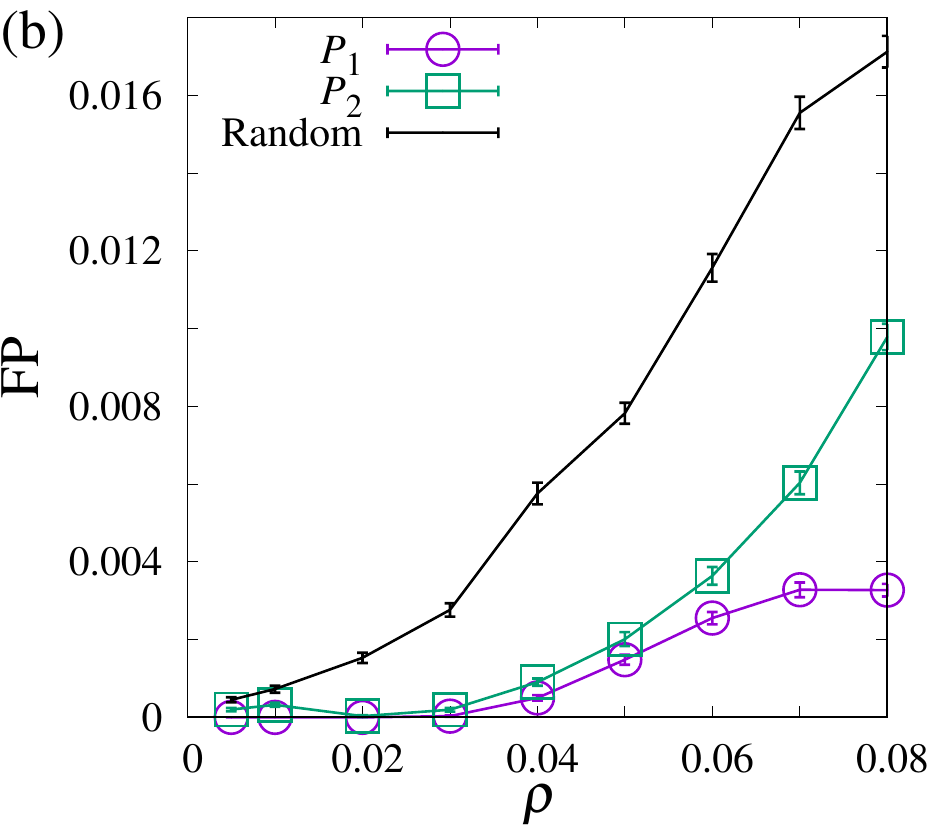}
\end{minipage}
    \caption{$\rho$-dependence of (a) true-positive and (b) false-positive  at $N=1000$ and $M=400$,
    where $M_{\mathrm{ini}}=300$ and $M_{\mathrm{ada}}=100$.
    In the initial stage, random pooling with $N_G=10$ is used.
    The error rates are fixed at $p_{\mathrm{TP
    }}=0.9$ and $p_{\mathrm{FP}}=0.05$.
    The horizontal dashed line in (a) represents $p_{\mathrm{TP}}$.
    For comparison, a random test with $M=400$ pools is shown.
    $P_1$ and $P_2$ denote ${\cal P}_1$ and ${\cal P}_2$ cases, respectively.}
    \label{fig:M_ini300}
\end{figure}


\Fref{fig:M_ini300} shows the-$\rho$-dependence of (a) TP and (b) FP at $N=1000$,
$M=400$ with $M_{\mathrm{ini}}=300$ and $M_{\mathrm{ada}}=100$.
The error probabilities are set at $p_{\mathrm{TP}}=0.9$ and 
$p_{\mathrm{FP}}=0.05$, and the group size in the initial stage is $N_G=10$.
$P_1$ and $P_2$ in the figure denote the 
results of the active pooling in the spaces ${\cal P}_1$ and ${\cal P}_2$,
respectively \footnote{
We note the heuristics used in the simulation.
When $\rho$ is sufficiently small such as $\rho=0.005$,
BP for the active pooling does not converge sometimes.
This is due to the overlapped pools; that is, certain pools are selected several times
in the adaptive stage.
It is known that rank deficiency can cause the instability of BP.
To avoid this problem, we exclude 
the already existing pool from the candidates in the subsequent stage
for the small-$\rho$ case.}.
For comparison, the results of 
random pooling are shown, where tests in $M_{\mathrm{ada}}$ steps 
are performed on random pools generated by the same rule
as the initial $M_{\mathrm{ini}}$-times tests.
Each data point represents the averaged value with respect to 
100 realizations of $\bm{Y}_{(M_{\mathrm{ini}})}$, $\bm{F}_{(M_{\mathrm{ini}})}$ and $\bm{X}^{(0)}$.
For any region of $\rho$, TP under a random test cannot exceed the
$p_{\mathrm{TP}}$, which is indicated by the horizontal line in \Fref{fig:M_ini300} (a).
The adaptive test improves TP and achieves $\mathrm{TP} >p_{\mathrm{TP}}$
when $\rho<0.02$ for ${\cal P}_1$ case and $\rho<0.04$ for 
${\cal P}_2$ case.
As shown in \Fref{fig:M_ini300} (b),
FP is smaller than $p_{\mathrm{FP}}$
even when the pooling is randomly determined,
but the adaptive test can further decrease FP.


\begin{figure}
\begin{minipage}{0.49\hsize}
\centering
    \includegraphics[width=1.7in]{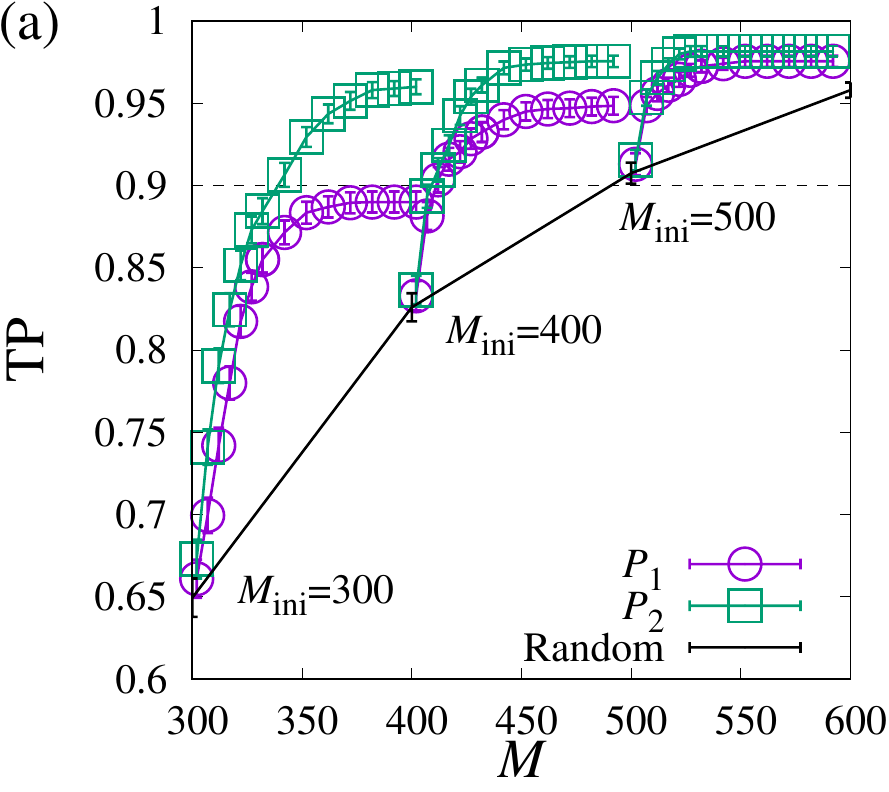}
\end{minipage}
\begin{minipage}{0.49\hsize}
\centering
\includegraphics[width=1.7in]{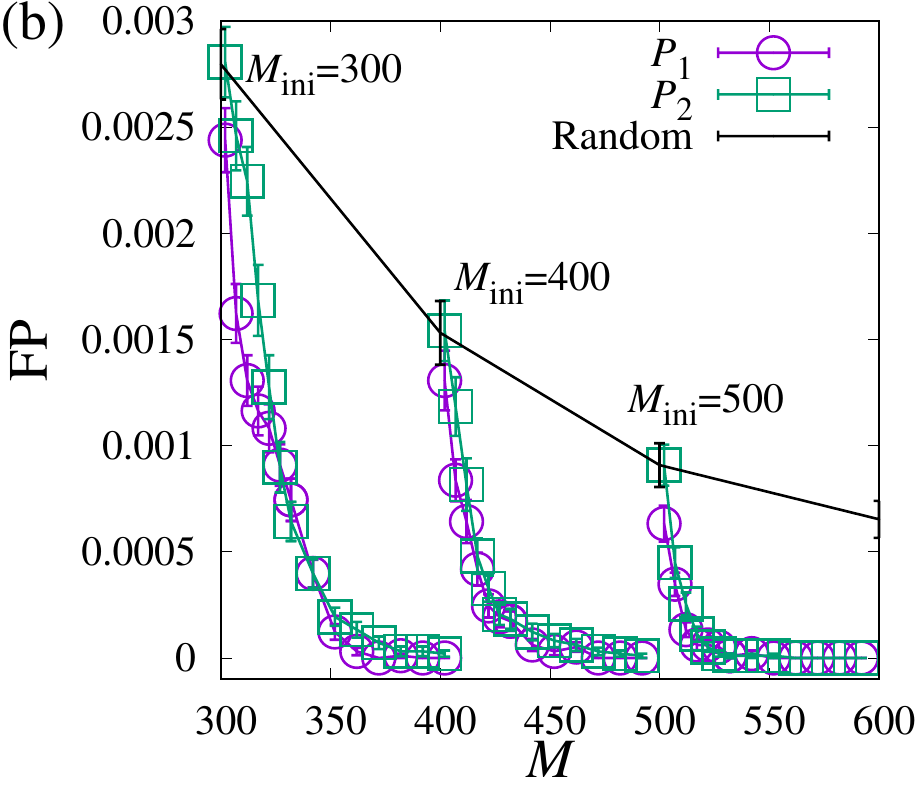}
\end{minipage}
    \caption{$M$-dependence of (a) true-positive and (b) false-positive 
    at $N=1000$ and $\rho=0.02$. 
    The adaptive tests are performed $100$ times
    after $M_{\mathrm{ini}}=300,~400$, and 500
    random tests with $N_G=10$.
    The error probabilities on the test are fixed at $p_{\mathrm{TP}}=0.9$ and $p_{\mathrm{FP}}=0.05$.
    The horizontal dashed line in (a) represents $p_{\mathrm{TP}}$.}
    \label{fig:rho_0d02}
\end{figure}

The performance of the adaptive test depends on the number of initial random tests
$M_{\mathrm{ini}}$.
\Fref{fig:rho_0d02} shows the $M_{\mathrm{ini}}$-dependence
of (a) TP and (b) FP at $N=1000$, $\rho=0.02$,
$p_{\mathrm{FP}}=0.9$, and $p_{\mathrm{TP}}=0.05$.
The pool size at the initial stage is $N_G=10$.
The figure also presents the results for $M_{\mathrm{ini}}=300$, 400, and 500.
The horizontal dashed line in (a) represents the TP probability of the test,
which is 0.9.
As $M_{\mathrm{ini}}$ increases, 
a high TP close to 1 is obtained via the adaptive test.
Moreover, for a large $M_{\mathrm{ini}}$, such as $M_{\mathrm{ini}}=500$,
the possible pooling space does not significantly influence the performance 
in terms of TP and FP.
Meanwhile, for a small $M_{\mathrm{ini}}$, the result of TP 
depends on the pooling space, and more accurate identification is achieved by ${\cal P}_2$.
The test results at the initial stage have large uncertainties when $M_{\mathrm{ini}}$ is small,
and hence, it is considered that 
the large pooling space is required for the effective sampling of the
uncertain pools.

As shown in \Fref{fig:rho_0d02} (a),
to achieve a high TP, the active pooling method requires
a smaller number of tests than that required by
the random pooling method.
For instance, the active pooling in ${\cal P}_2$ 
with $M_{\mathrm{ini}}=300$ initial stage
results in a high $\mathrm{TP}>p_{\mathrm{TP}}$
after the $M_{\mathrm{ada}}=40$ adaptive stage,
namely $M=340$.
Meanwhile, the random pooling
achieves $\mathrm{TP}>p_{\mathrm{TP}}$
with almost $M=500$ tests.
With regard to the improvement of TP,
the adaptive method helps effectively 
identify infected patients
with a small number of tests.

\begin{figure}
\begin{minipage}{0.49\hsize}
\centering
    \includegraphics[width=1.7in]{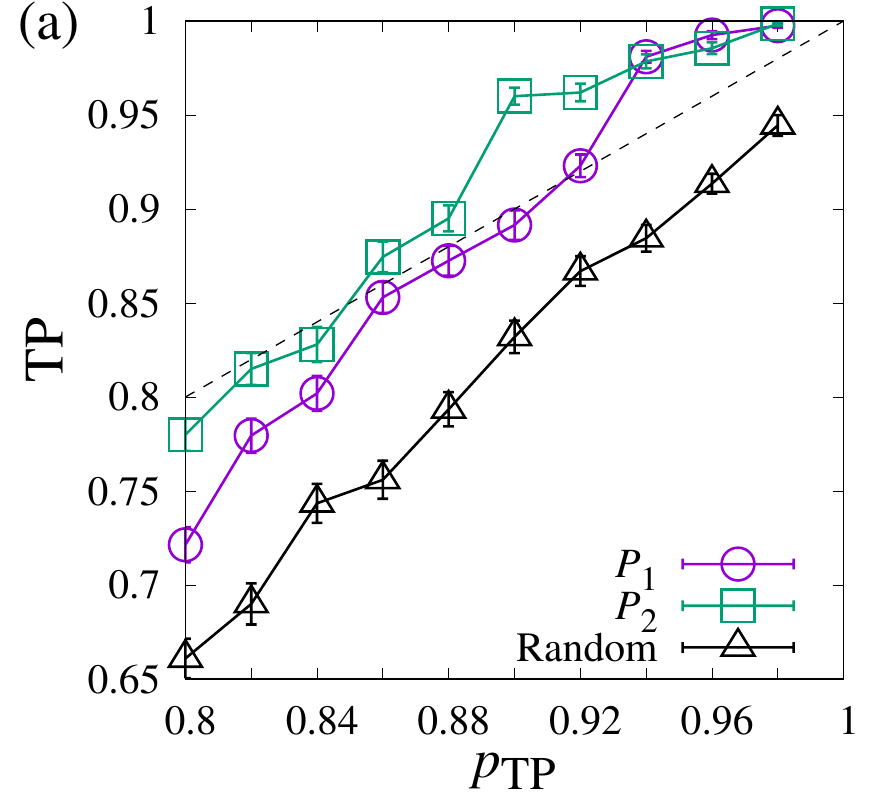}
\end{minipage}
\begin{minipage}{0.49\hsize}
\centering
\includegraphics[width=1.7in]{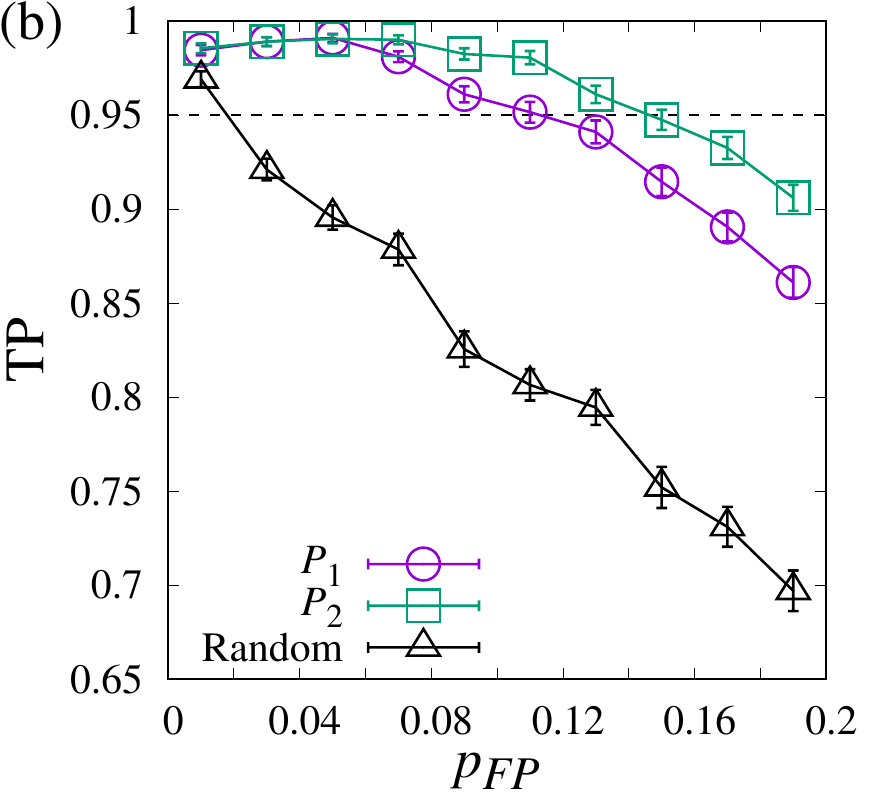}
\end{minipage}
    \caption{$p_{\mathrm{TP}}$-dependence of true-positive
    for (a) $p_{\mathrm{FP}}=0.05$ and (b) $p_{\mathrm{TP}}=0.95$ 
    at $N=1000$ and $\rho=0.02$. 
    The adaptive tests are conducted $100$ times
    after $M_{\mathrm{ini}}=300$; 
    hence, $M=400$ in total.
    The pool size in the random tests is $N_G=10$.
    The horizontal dashed lines represent $\mathrm{TP}=p_{\mathrm{TP}}$.}
    \label{fig:rho_0d02_err}
\end{figure}

The active pooling is robust to the errors in the test, 
compared with the random pooling.
\Fref{fig:rho_0d02_err} shows (a) the $p_{\mathrm{TP}}$-dependence of TP
for $p_{\mathrm{FP}}=0.05$
and (b) the $p_{\mathrm{FP}}$-dependence of TP for $p_{\mathrm{TP}}=0.95$
at $N=1000$, $M_{\mathrm{ini}}=300$, $M_{\mathrm{ada}}=100$, and $\rho=0.02$.
The random tests in the initial stage are performed on pools of size $N_G=10$.
For the random pooling case,
$\mathrm{TP}>p_{\mathrm{TP}}$ is achieved only when
$p_{\mathrm{FP}}$ is sufficiently small such as $p_{\mathrm{FP}}<0.02$.
The adaptive test improves TP,
and the parameter region where $\mathrm{TP}>p_{\mathrm{TP}}$
is extended in particular for the case ${\cal P}_2$. 

These results indicate the efficiency of the 
active pooling design based on predictive distribution
in group testing.
However, a limitation of this approach is that the computational cost involved is higher than that of the
non-adaptive approach.
The estimation of the infection probability
via the BP algorithm $M_{\mathrm{ada}}$ steps is obtained again;
hence, the computational cost
of the adaptive approach is approximately $M_{\mathrm{ada}}$ times larger than that of the non-adaptive approach.
However, the adaptive approach achieves accurate estimation
using a small number of tests.
The trade-off between the reduction of the operating cost involved in tests
and the increase in the computation time of inference 
should be considered for practically apply the
adaptive approach.



\section{Summary and discussion}

In this paper, we propose an active pooling design in
adaptive group testing,
where the pool for the subsequent stage
is determined based on the Bayesian posterior predictive distribution
under the test outcomes in the previous stage.
The proposed method was implemented using the BP algorithm,
and the identification of infected patients
using adaptive tests is demonstrated to be more accurate than that using randomly designed pools.
In particular, the active pooling design reduced
the number of required tests to achieve $\mathrm{TP}>p_{\mathrm{TP}}$.
Further, the proposed method is robust to test errors 
and $\mathrm{TP}>p_{\mathrm{TP}}$ holds in
smaller $p_{\mathrm{TP}}$ and larger $p_{\mathrm{FP}}$, 
compared with the approach that uses randomly designed pools.

In the current study, we restrict the possible pooling space
within ${\cal P}_1$ and ${\cal P}_2$.
Mathematically, 
more uncertain pool can be considered removing this restriction,
and further improvement in the TP and FP rates is expected.
However, the straightforward calculation of 
the predictive entropy for all possible $\pi$
is computationally intractable.
Hence, some approximation will be required.
An efficient sampling method in $\pi\in{\cal P}$
to find the uncertain pool should be developed
such as the Markov chain Monte Carlo method.

We focused on the MAP estimator to convert the 
estimated infection probability, which is $[0,1]$ variable,
into the state of patients, $\{0,1\}$
variable, because of its simplicity;
however, changing the decision threshold 
from 0.5 results in improvements in the TP rate.
For example, the estimate using the confidence interval constructed based on the 
bootstrap method was obtained. Further, the TP rate obtained using this method is higher than that obtained using the MAP estimator \cite{Sakata}; however, its computational cost is unrealistic to accompany the 
active pooling procedure.
The receiver operating characteristic (ROC) analysis 
is a promising method to understand the appropriate decision threshold \cite{ROC1,ROC2}.
Along with the ROC analysis,
the mathematical background of the active pooling 
proposed in this paper is expected to be established.


\begin{acknowledgments}

This work was accomplished thanks to the author’s pleasant discussions with Yukito Iba.
Further, the author thanks Koji Hukushima, Yoshiyuki Kabashima, and Satoshi Takabe
for their helpful comments and discussions.
This research was partially supported by Grant-in-Aid for Scientific Research 19K20363 from the Japanese Society for the Promotion of Science (JSPS) and JST PRESTO Grant Number JPMJPR19M2, Japan.

\end{acknowledgments}

\appendix

\section{BP algorithm for group testing}
\label{sec:app_BP}

We denote $\pi(\mu)$ and ${\cal G}(i)$
as the indices of the patients in the $\mu$-th pool
and those of the pools in which the $i$-th patient is included,
respectively.
For the edge that connects the $\mu$-th factor (test)
and the $i$-th variable (patient), two types of messages 
$m_{i\to \mu}(X_i)$ and $\tilde{m}_{\mu\to i}(X_i)$ are defined.
Intuitively, the 
messages $m_{i\to \mu}(X_i)$ and $\tilde{m}_{\mu\to i}(X_i)$
represent the marginal distributions of $X_i$
before and after the $\mu$-th test is performed,
respectively.
The variable $X_i$ is binary. Hence, the messages can be expressed by
the Bernoulli probability $\theta_{i\to\mu}$ and $\tilde{\theta}_{\mu\to i}$ given by 
\begin{align}
    \theta_{i\to\mu}=\frac{\rho\prod_{\nu\in{\cal G}(i)\backslash \mu} \tilde{\theta}_{\nu\to i}}{Z_{i\to\mu}},~~~
        \tilde{\theta}_{\mu\to i}=\frac{U_\mu}{\tilde{Z}_{\mu\to i}}
\end{align}
where $U_\mu=p_{\mathrm{TP}}Y_\mu+(1-p_{\mathrm{TP}})(1-Y_\mu)$,
$W_\mu=p_{\mathrm{FP}}Y_\mu+(1-p_{\mathrm{FP}})(1-Y_\mu)$ and
\begin{align}
    \tilde{Z}_{\mu\to i}&=U_\mu\left(2-\prod_{j\in\pi(\mu)\backslash i} (1-\theta_{j\to\mu})\right)
    +W_\mu\prod_{j\in\pi(\mu)\backslash i}(1-\theta_{j\to\mu})\\
    Z_{i\to\mu}&=\rho\prod_{\nu\in{\cal G}(i)\backslash \mu} \tilde{\theta}_{\nu\to i}+
    (1-\rho)\prod_{\nu\in{\cal G}(i)\backslash \mu} (1-\tilde{\theta}_{\nu\to i}).
\end{align}
The BP algorithm consists of the recursive update of $\theta_{i\to\mu}$ and 
$\tilde{\theta}_{\mu\to i}$, and 
at the fixed point, the infection probability is given by \cite{Johnson,Mezard_GT,Sakata}
\begin{align}
    \hat{\theta}_i=\frac{\rho\prod_{\mu\in{\cal G}(i)}\tilde{\theta}_{\mu\to i}}
    {\rho\prod_{\mu\in{\cal G}(i)}\tilde{\theta}_{\mu\to i}
    +(1-\rho)\prod_{\mu\in{\cal G}(i)}(1-\tilde{\theta}_{\mu\to i})}.
    \label{eq:infect_prob}
\end{align}

\section{Calculation of susceptibility using the BP algorithm}
\label{sec:app_correlation}

\begin{figure}
\begin{minipage}{0.49\hsize}
\centering
    \includegraphics[width=1.65in]{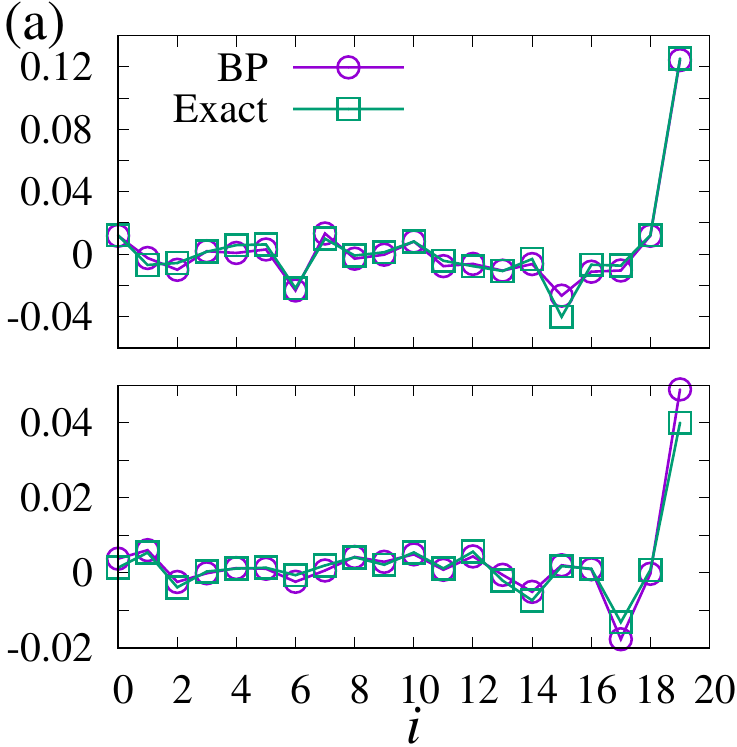}
\end{minipage}
\begin{minipage}{0.49\hsize}
\centering
\includegraphics[width=1.75in]{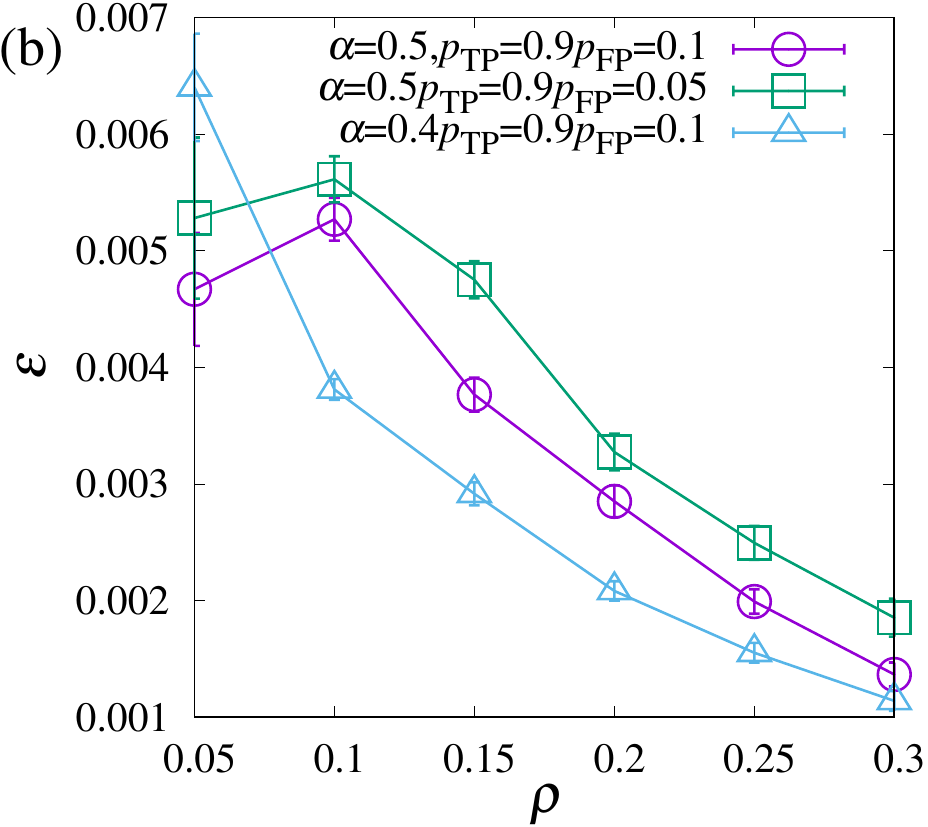}
\end{minipage}
    \caption{(a) Examples of susceptibility calculated using the BP algorithm and exact computation
    at $N=20$, $M=10$, $N_G=10$, $N_O=5, \rho=0.1$, $p_{\mathrm{TP}}=0.95$, and $p_{\mathrm{FP}}=0.05$.
    $\chi_{ij}$ for $j=20$ is shown for two different realizations of $\bm{Y},\bm{F}$, and $\bm{X}^{(0)}$.
    (b) Quantification of the difference between susceptibilities given by the BP algorithm
    and the exact calculations at $N=20$ and $N_G=10$ for several parameters using $\epsilon$. }
    \label{fig:chi_comparison}
\end{figure}

\begin{figure}
\begin{minipage}{0.49\hsize}
\centering
    \includegraphics[width=1.7in]{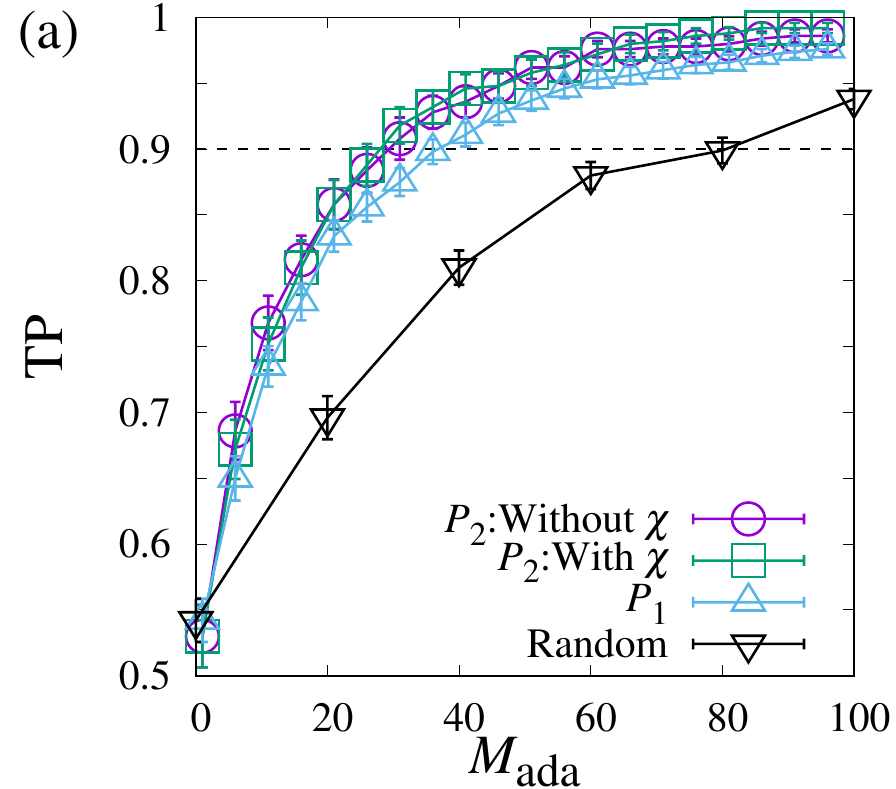}
\end{minipage}
\begin{minipage}{0.49\hsize}
\centering
\includegraphics[width=1.7in]{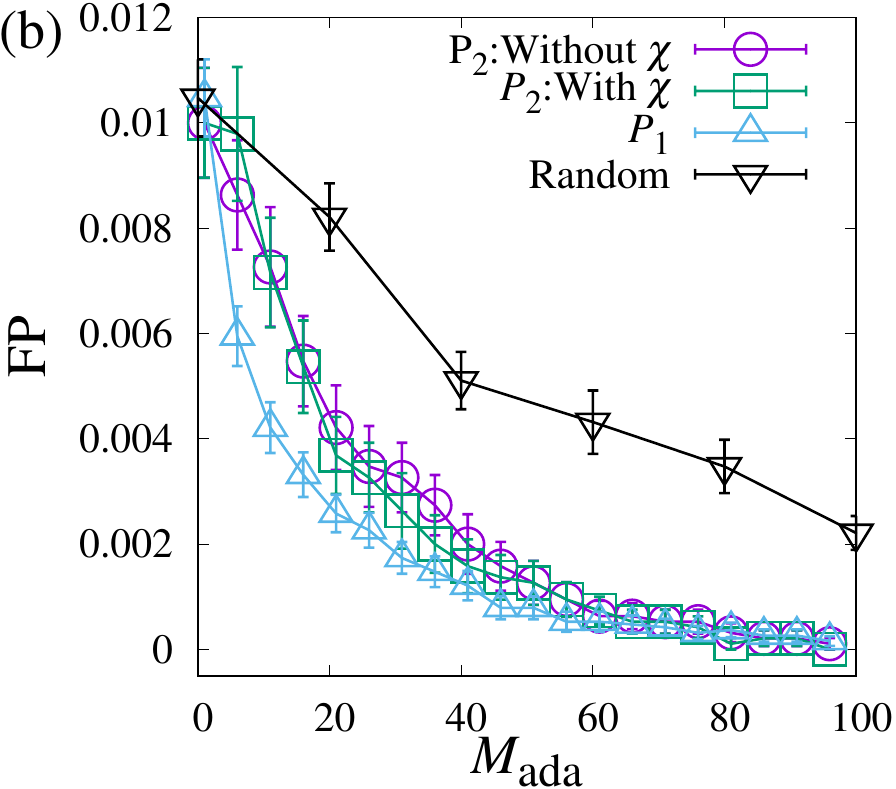}
\end{minipage}
\caption{Comparison of (a) true-positive and (b) false-positive 
for ${\cal P}_2$ case considering the susceptibility,
${\cal P}_2$ case without considering the susceptibility,
and ${\cal P}_1$ case, 
at $N=200$, $M_{\mathrm{ini}}=80$,
$\rho=0.05$, $p_{\mathrm{TP}}=0.9$, and $p_{\mathrm{FP}}=0.1$. 
The tests in the initial stage are performed on the randomly designed pools 
with $N_G=10$ and $N_O=4$.}
\label{fig:TPFP_chi}
\end{figure}

A widely used method to calculate susceptibility
in the framework of the BP algorithm
is susceptibility propagation  \cite{Mezard-Mora,Yasuda-Tanaka},
where recursive update of tensors
that give susceptibility is introduced
on the basis of linear-response theory.
In the current problem setting,
the variables to be estimated obey
the Bernoulli probability.
Hence, we can compute the susceptibility in a simpler way.

Let us denote the expectation of $X_j$ under the posterior conditional distribution
$\sum_{\bm{X}\backslash i,j}P_{\mathrm{post}}(X_j,X_i=1,\bm{X}_{\backslash i,j}|\bm{Y})$
as $\theta_j^{|X_i=1}\equiv E_{\mathrm{post}|\bm{Y}}[X_j|X_i=1]$ $(i\neq j)$.
This expectation value is evaluated using the BP algorithm
by fixing $\theta_{i\to\mu}=1$ and $\tilde{\theta}_{\mu\to i}=1$
for $\mu\in{\cal G}(i)$.
The conditional expectation value obtained using the BP algorithm
is denoted as $\hat{\theta}_j^{|X_i=1}$. Thus, the susceptibility is 
given by $\hat{\chi}_{ij}=\hat{\theta}_i\hat{\theta}_j^{|X_i=1}-\hat{\theta}_i\hat{\theta}_j$.
We can show that the symmetry
$\hat{\theta}_i\hat{\theta}_j^{|X_i=1}=\hat{\theta}_i^{|X_j=1}\hat{\theta}_j$ holds.

To check the accuracy of the susceptibility derived using the BP algorithm,
we compute the exact posterior distribution
by sampling all configurations in $\{0,1\}^N$.
Examples of the exact susceptibility and the approximated one
are shown in \Fref{fig:chi_comparison}(a) at $N=20$, $M=10$, 
$\rho=0.1$, $p_{\mathrm{TP}}=0.95$, and $p_{\mathrm{FP}}=0.05$,
where the $i$-dependence of $\chi_{i,20}$ is shown for 
two different realizations of $\bm{Y},\bm{F}$, and $\bm{X}^{(0)}$.
Here, the pooling matrix is randomly generated 
to be $N_G=10$ and $N_O=5$.
The difference between $\chi$ and $\hat{\chi}$ 
is quantified by $\epsilon\equiv\sum_{i<j}(\chi_{ij}-\hat{\chi}_{ij})^2/ \{N(N-1)/ 2\}$,
whose behavior is shown in \Fref{fig:chi_comparison} (b) at $N=20$ and 
different values of $\alpha\equiv M\slash N$, $p_{\mathrm{TP}}$, and $p_{\mathrm{FP}}$.
For any parameter region, $\epsilon$ is $O(10^{-3})$.
Therefore, we consider that the BP algorithm provides a reasonable approximation of the susceptibility
and expect that it is also applicable for larger $N$.

In \Fref{fig:TPFP_chi}, (a) TP and (b) FP are shown 
for the cases when the susceptibility is considered (denoted by `$P_2$: with $\chi$') and not considered
(denoted by `$P_2$: without $\chi$'); 
namely, \eref{eq:q_pi2} is used to determine the pool in the subsequent stage
by substituting $\hat{\chi}$ calculated by BP into $\chi$
at $N=200$, $\rho=0.05$, $p_{\mathrm{TP}}=0.9$, and $p_{\mathrm{FP}}=0.1$. 
Each data point is averaged over 50 samples of $\bm{F}$, $\bm{X}^{(0)}$, and $\bm{Y}$.
The initial stage consists of $M_{\mathrm{ini}}=80$ random tests 
with $N_G=10$ and $N_O=4$.
The ${\cal P}_1$ case is compared with 
the random case with the same test at the initial stage.
Considering the susceptibility,
a slight improvement in TP is observed.

Following the same procedure, we can compute a higher order correlation in principle.
For example, $E_{\mathrm{post}|\bm{Y}}[X_i|X_j=1,X_k=1]$
is obtained by fixing $\theta_{j\to\mu}=1$, $\theta_{k\to\nu}=1$, and 
$\tilde{\theta}_{\mu\to j}=1$, $\tilde{\theta}_{\nu\to k}=1$
for $\mu\in{\cal G}(j)$ and $\nu\in{\cal G}(k)$.

\providecommand{\noopsort}[1]{}\providecommand{\singleletter}[1]{#1}%

\end{document}